\documentclass[11pt,a4paper]{article}
\usepackage{naaclhlt2018}
\usepackage{times}
\usepackage{latexsym}
\usepackage{paralist}
\usepackage{amsmath}
\usepackage{subfig}
\usepackage{floatrow}
\usepackage{contour}
\usepackage[normalem]{ulem}

\contourlength{0.8pt}

\newcommand{\myuline}[1]{%
  \uline{\phantom{#1}}%
  \llap{\contour{white}{#1}}%
}
\usepackage{url}
\usepackage{booktabs}

\usepackage[utf8]{inputenc}			% кодировка исходного текста
\usepackage[russian,english]{babel}	% локализация и переносы

\usepackage{linguex}
\renewcommand{\firstrefdash}{} 
\renewcommand{\firstrefdash}{} 
\usepackage[]{tikz-dependency}
\usepackage[colorinlistoftodos,prependcaption,textsize=scriptsize]{todonotes}

\aclfinalcopy % Uncomment this line for the final submission

\title{Colorless green recurrent networks dream hierarchically}

\author{Kristina Gulordava\footnotemark{} \\
  Department of Linguistics \\
  University of Geneva \\
  {\tt \small kristina.gulordava@unige.ch}
  \And
  Piotr Bojanowski \\
  Facebook AI Research \\
  Paris\\
  {\tt \small bojanowski@fb.com} \\ 
   \And
  Edouard Grave \\
  Facebook AI Research \\
  New York \\
  {\tt \small egrave@fb.com} 
  \AND
  Tal Linzen\\
  Department of Cognitive Science\\
  Johns Hopkins University \\
  {\tt \small tal.linzen@jhu.edu} \\
  \\\And
    Marco Baroni \\
  Facebook AI Research \\
  Paris \\
  {\tt \small mbaroni@fb.com} \\}

\date{}

\begin{document}
\maketitle

\begin{abstract}
  Recurrent neural networks (RNNs) have achieved impressive results in
  a variety of linguistic processing tasks, suggesting that they can
  induce non-trivial properties of language. We
  investigate here to what extent RNNs learn to track abstract hierarchical
  syntactic structure. We test whether RNNs trained with a generic
  language modeling objective in four languages (Italian, English,
  Hebrew, Russian) can predict long-distance number agreement in
  various constructions. We include in our evaluation
  nonsensical sentences where RNNs cannot rely on semantic or lexical
  cues (``The colorless green \myuline{ideas} I ate with the chair
  \myuline{sleep} furiously''), and, for Italian, we compare model
  performance to human intuitions. Our language-model-trained RNNs
  make reliable predictions about long-distance
  agreement, and do not lag much behind human performance. We thus
  bring support to the hypothesis that RNNs are not just
  shallow-pattern extractors, but they also acquire deeper grammatical
  competence.
\end{abstract}

\setlength{\pdfpxdimen}{1in/350}

\renewcommand{\thefootnote}{\fnsymbol{footnote}}

\footnotetext[1]{The work was conducted during the internship at Facebook AI Research, Paris.}
\renewcommand*{\thefootnote}{\arabic{footnote}}

\section{Introduction}

Recurrent neural networks (RNNs; \citealp{Elman:1990}) are general sequence
processing devices that do not explicitly encode the hierarchical structure
that is thought to be essential to natural language \citep{Everaert:etal:2015}.
Early work using artificial languages showed that they may nevertheless be able
to approximate context-free languages \citep{Elman:1991}. More recently, RNNs
have achieved impressive results in large-scale tasks such as
language modeling for speech recognition and machine translation, and are by now
standard tools for sequential natural language tasks
\citep[e.g.,][]{Mikolov:etal:2010,Graves:2012,Wu:etal:2016}. This suggests that
RNNs may learn to track grammatical structure even when trained on noisier
natural data. The conjecture is supported by the success of RNNs as feature
extractors for syntactic parsing
\citep[e.g.,][]{Cross:Huang:2016,Kiperwasser:Goldberg:2016,Zhang:etal:2017}.

\newcite{Linzen:etal:2016} directly evaluated the extent to which RNNs can
approximate hierarchical structure in corpus-extracted natural language data. 
They tested whether RNNs can learn to predict English subject-verb agreement, a
task thought to require hierarchical structure in the general case (``the
\myuline{girl} the boys like\ldots{} \myuline{is} or \myuline{are}?'').
Their experiments confirmed that RNNs can, in principle, handle such constructions.
However, in their study RNNs could only succeed when provided with explicit
supervision on the target task. Linzen and colleagues argued that the unsupervised language
modeling objective is not sufficient for RNNs to induce the syntactic
knowledge necessary to cope with long-distance agreement.

The current paper reevaluates these conclusions. We strengthen the
evaluation paradigm of Linzen and colleagues in several ways. Most importantly, their analysis did not rule out the possibility that RNNs might be relying on \emph{semantic} or
\emph{collocational/frequency-based} information, rather than purely on
syntactic structure. In ``\myuline{dogs} in the neighbourhood often
\myuline{bark}'', an RNN might get the right agreement by encoding information
about what typically barks (dogs, not neighbourhoods), without relying on more
abstract structural cues. In a follow-up study to Linzen and colleagues', \citet{Bernardy:Lappin:2017}
observed that RNNs are better at long-distance agreement when they construct
rich lexical representations of words, which suggests effects of this sort might
indeed be at play.

We introduce a method to probe the syntactic abilities of RNNs that abstracts
away from potential lexical, semantic and frequency-based confounds. Inspired by
Chomsky's \shortcite{Chomsky:1957} insight that ``grammaticalness cannot be
identified with meaningfulness'' (p.~106), we test long-distance agreement both
in standard corpus-extracted examples and in comparable nonce sentences that are
grammatical but completely meaningless, e.g., (paraphrasing Chomsky): ``The
colorless green \myuline{ideas} I ate with the chair \myuline{sleep}
furiously''.

We extend the previous work in three additional ways. First,
alongside English, which has few morphological cues to agreement, we examine
Italian, Hebrew and Russian, which have richer  morphological systems. Second,
we go beyond subject-verb agreement and develop an automated method to harvest a
variety of long-distance number agreement constructions from treebanks. Finally,
for Italian, we collect human judgments for the tested sentences, providing an
important comparison point for RNN performance.\footnote{The code to reproduce our experiments and the data used for training and evaluation, including the human judgments in Italian, can be found at \url{https://github.com/facebookresearch/colorlessgreenRNNs}.}

We focus on the more
interesting unsupervised setup, where RNNs are trained to perform generic,
large-scale language modeling (LM): they are not given explicit evidence, at
training time, that they must focus on long-distance agreement, but they are
rather required to track a multitude of cues that might help with word
prediction in general.

Our results are encouraging.  RNNs trained with a LM objective solve
the long-distance agreement problem well, even on nonce sentences.
The pattern is consistent across languages, and, crucially, not far
from human performance in Italian.  Moreover, RNN performance on
language modeling (measured in terms of perplexity) is a good
predictor of long-distance agreement accuracy. This suggests that the
ability to capture structural generalizations is an important aspect
of what makes the best RNN architectures so good at language modeling.
Since our positive results contradict, to some extent, those of
\citet{Linzen:etal:2016}, we also replicate their relevant experiment
using our best RNN (an LSTM). We outperform their models, suggesting
that a careful architecture/hyperparameter search is crucial to obtain
RNNs that are not only good at language modeling, but able to extract
syntactic generalizations.

\section{Constructing a long-distance agreement benchmark}
\label{sec:long_distance}

\paragraph{Overview.} We construct our number agreement test sets as
follows. \textbf{Original} sentences are automatically extracted from
a dependency treebank. They are then converted into \textbf{nonce}
sentences by substituting all content words with random words 
with the same morphology, resulting in grammatical but nonsensical
sequences. An LM is evaluated on its predictions for the target
(second) word in the dependency, in both the original and
nonce sentences.

\paragraph{Long-distance agreement constructions.}

Agreement relations, such as subject-verb agreement in English, are an ideal test bed for the syntactic abilities of LMs,
because the form of the second item (the target) is
predictable from the first item (the cue). Crucially,
the cue and the target are linked by a \emph{structural} relation, where
linear order in the word sequence does not matter
\cite{Everaert:etal:2015}. Consider the following subject-verb
agreement examples: ``the \myuline{girl} \myuline{thinks}\ldots'', ``the
\myuline{girl} [you met] \myuline{thinks}\ldots'', ``the \myuline{girl}
[you met yesterday] \myuline{thinks}\ldots'', ``the \myuline{girl} [you
met yesterday through her friends] \myuline{thinks}\ldots''. In all
these cases, the number of the main verb ``thinks'' is determined by
its subject (``girl''), and this relation depends on the syntactic structure of
the sentence, not on the linear sequence of words. As the last
sentence shows, the word directly preceding the verb can even be a noun
with the opposite number (``friends''), but this does not influence
the structurally-determined form of the verb.

\begin{figure*}
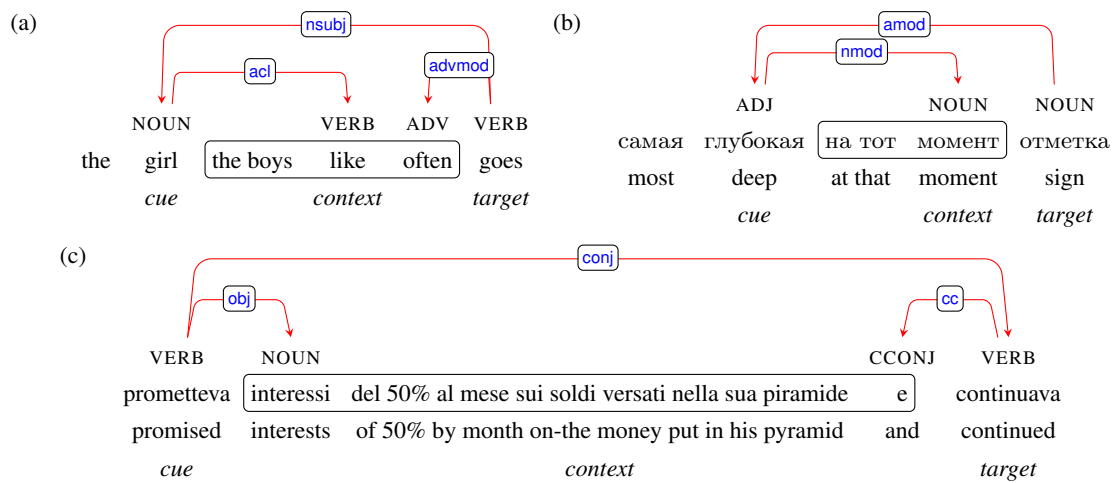

\centering
\sidesubfloat[]{
    \label{f:example:english}
\begin{dependency}[edge slant=1pt,edge style=red, label style = {text = blue,font=\small\sffamily}]
\begin{deptext}[font=\small,column sep=0.25em, row sep=.1ex]
    \& \textsc{noun} \& \& \textsc{verb} \& \textsc{adv} \& \textsc{verb} \\
the \& girl \& the boys \& like \& often \& goes \\ 
\& \textit{cue} \& \& \textit{context} \& \& \textit{target} \\
\end{deptext}
   \depedge[edge unit distance=1.2ex]{2}{4}{acl}
   \depedge[edge unit distance=1.5ex]{6}{2}{nsubj}
   \depedge{6}{5}{advmod}
   \wordgroup{2}{3}{5}{relc}
\end{dependency}
}
\sidesubfloat[]{
    \label{f:example:russian}
\begin{dependency}[edge slant=1pt,edge style=red, label style={text=blue, font=\small\sffamily}]
\begin{deptext}[font=\small, column sep=0.25em, row sep=.1ex]
    \& \textsc{adj} \& \&  \textsc{noun} \& \textsc{noun} \\
\foreignlanguage{russian}{самая} \& \foreignlanguage{russian}{глубокая} \& \foreignlanguage{russian}{на тот} \& \foreignlanguage{russian}{момент} \& \foreignlanguage{russian}{отметка}  \\ 
    most \& deep \& at that \& moment \& sign \\
    \& \textit{cue} \& \& \textit{context} \& \textit{target} \\
    \end{deptext}
   \depedge[edge unit distance=1.2ex]{2}{4}{nmod}
   \depedge[edge unit distance=1.5ex]{5}{2}{amod}   
   \wordgroup{2}{3}{4}{relc}
   
\end{dependency}
}

\sidesubfloat[]{
    \label{f:example:italian}
\begin{dependency}[edge style=red, label style = {text = blue,font=\small\sffamily}]
   \begin{deptext}[font=\small, column sep=0.25em, row sep=.1ex]
       \textsc{verb} \& \textsc{noun} \& \& \textsc{cconj} \& \textsc{verb} \\
prometteva \& interessi \& del 50\% al mese sui soldi versati nella sua piramide \& e \& continuava
  \\ 
    promised \& interests \& of 50\% by month on-the money put in his pyramid \& and \& continued \\
    \textit{cue} \& \& \textit{context} \& \& \textit{target} \\
    \end{deptext}
   \depedge{1}{2}{obj}
   \depedge{5}{4}{cc}
   \depedge[edge unit distance=1.5ex]{1}{5}{conj}
   \wordgroup{2}{2}{4}{relc}
\end{dependency}
}
\caption{Example agreement constructions defined by a dependency and the separating context, in (a)~English, (b) Russian and (c) Italian.}
\label{f:agreement_constructions}
\end{figure*}

When the cue and the target are adjacent (``the \myuline{girl}
\myuline{thinks}\ldots''), an LM can predict the target without access
to syntactic structure: it can simply extract the relevant morphosyntactic features of words (e.g., number) and record the co-occurrence frequencies of patterns such as N$_{Plur}$ V$_{Plur}$
\cite{Mikolov:etal:2013a}. Thus, we focus here on \emph{long-distance}
agreement, where an arbitrary number of words can occur between the
elements of the agreement relation.  We limit ourselves to \emph{number}
agreement (plural or singular), as it is the only
overt agreement feature shared by all of the languages we study.

\paragraph{Identifying candidate constructions.}
We started by collecting pairs
of part-of-speech (POS) tags connected by a dependency arc. Independently of
which element is the head of the relation, we refer to the first item as the \textbf{cue} and to the second as the \textbf{target}. We additionally refer to
  the POS sequence characterizing the entire pattern as a
  \textbf{construction}, and to the elements in the middle as
  \textbf{context}.

For each candidate construction, we
collected all of the contexts 
in the corpus that intervene between the cue and the target (we define contexts as the sequence of POS tags of the top-level nodes in the dependency subtrees).  For example, for the English subject-verb
agreement construction shown in Fig.~\ref{f:example:english}, the context is
defined by \textsc{verb} (head of the relative clause) and \textsc{adv} (adverbial
modifier of the target verb), which together dominate the 
sequence ``the boys like often''. For the Russian adjective-noun
agreement construction in Fig.~\ref{f:example:russian}, the context is \textsc{noun}, because
in the dependency grammar we use the noun ``moment'' is the
head of the prepositional phrase ``at that moment'', which modifies the
adjective ``deep''. The candidate agreement pair and the context form
a construction, which is characterized by a sequence of POS tags, e.g., \textsc{noun
verb adv verb} or \textsc{verb noun cconj verb} (Fig.~\ref{f:example:italian}).

Our constructions do not necessarily correspond to standard syntactic
structures. The English subject-verb agreement construction \textsc{noun verb verb},
for example, matches both object and subject relative clause contexts,
e.g., ``\myuline{girl} the boys like \myuline{is}'' and ``\myuline{girls} who stayed at home
\myuline{were}''. Conversely, standard syntactic structures might be
split between different constructions, e.g., relative clause contexts
occur in both \textsc{noun verb verb} and \textsc{noun verb adv verb}
constructions (the latter is illustrated by the English example in
Fig.~\ref{f:example:english}). %

Construction contexts can contain a variable numbers of
words. Since we are interested in challenging cases, we only considered
cases in which at least three tokens intervened between the cue and the target.

\paragraph{Excluding non-agreement constructions.} In the next step, we excluded constructions in which the candidate cue and target did not agree in number in all of the instances of the construction in the treebank (if both the cue and the target were morphologically annotated for number). This step retained English subject-verb constructions, for example, but excluded verb-object
constructions, since any form of a verb can appear both with singular
and plural objects. To focus on robust agreement patterns, we only
kept constructions with at least 10 instances of both plural and
singular agreement.

When applied to the treebanks we used (see Section~\ref{sec:exp_setup}), this step resulted in between two (English) and 21
(Russian) constructions per language. English has the poorest
morphology and consequently the lowest number of patterns with
identifiable morphological agreement. Only the VP-conjunction
construction (Fig.~\ref{f:example:italian}) was
identified in all four languages. Subject-verb agreement constructions
were extracted in all languages but Russian; Russian has relatively
flexible word order and a noun dependent preceding a head verb is not
necessarily its subject. The full list of extracted constructions in English and Italian is
given in Tables~\ref{t:two_hard_constructions} and \ref{t:italian_mturk}, respectively. For the other languages, see the Supplementary Material (SM).\footnote{The SM is available as a standalone file on the project's public repository.}

\paragraph{Original sentence test set.} Our ``original'' sentence test set included all sentences from each construction
where all words from the cue and up to and including the target occurred
in the LM vocabulary (Section
\ref{sec:exp_setup}), and where the singular/plural
counterpart of the target occurred in the treebank and in the language
model vocabulary (this is required by the evaluation procedure
outlined below).  The total counts of constructions and original
sentences in our test sets are provided in Table~\ref{t:all_langs}. The average number of context words separating the cue
and the target ranged from 3.6 (Hebrew) to 4.5 (Italian).

\paragraph{Generating nonce sentences.}
We generated nine nonce variants of each original sentence as
follows. Each content word (noun, verb, adjective, proper noun,
numeral, adverb) in the sentence was substituted by another random
content word from the treebank with matching POS and morphological
features. To avoid forms that are ambiguous between several POS, which are particularly frequent in English (e.g., plural noun and singular verb forms), we   excluded the forms that appeared with a different POS more than 10\% of the time in the treebank. 
Function words (determiners, pronouns, adpositions, particles) and
punctuation were left intact. 
For example, we generated the nonce \ref{ex:nonce} from the original sentence \ref{ex:original}:

\ex.\a.\label{ex:original}It \myuline{presents} the case for marriage equality and \myuline{states}\ldots
\b.\label{ex:nonce}It \myuline{stays} the shuttle for honesty insurance and \myuline{finds}\ldots

Note that our generation procedure is
based on morphological features and does not guarantee that argument
structure constraints are respected (e.g.,
``it stays the shuttle'' in \ref{ex:nonce}).

\paragraph{Evaluation procedure.}

For each sentence in our test set, we retrieved from our treebank the
form that is identical to the agreement target in all morphological
features except number (e.g., ``finds'' instead of ``find'' in
\ref{ex:nonce}).  Given a sentence with prefix $p$ up to and excluding
the target, we then compute the probabilities $P(t_1 | p)$ and
$P(t_2 | p)$ for the singular and plural variants of the target, $t_1$ and $t_2$, based
on the language model. Following \newcite{Linzen:etal:2016}, we say
that the model identified the correct target if it assigned a higher
probability to the form with the correct number. In \ref{ex:nonce}, for 
example, the model should assign a higher probability to ``finds''
than ``find''.\footnote{Obviously, in the nonce cases, the LMs never
  assigned the highest overall probability to either of the two
  candidates. Qualitatively, in such cases LMs assigned the largest
  absolute probabilities to plausible frequent words.}

\section{Experimental setup}
\label{sec:exp_setup}

\paragraph{Treebanks.}
We extracted our test sets from the Italian, English, Hebrew and Russian
Universal Dependency treebanks (UD, v2.0,
\citealp{Nivre:2016}). The English and Hebrew treebanks were post-processed to obtain a richer morphological annotation at the word level (see SM for details).

\paragraph{LM training data.}
Training data for Italian, English and Russian
were extracted from the respective Wikipedias. We
downloaded recent dumps, extracted the raw text from them using
WikiExtractor\footnote{\url{https://github.com/attardi/wikiextractor}}
and tokenized it with TreeTagger \cite{Schmid:1995}. We also used the
TreeTagger lemma annotation to filter out sentences with more than 5\%
unknown words. For Hebrew, we used the preprocessed Wikipedia corpus
made available by Yoav
Goldberg.\footnote{\url{http://u.cs.biu.ac.il/~yogo/hebwiki/}} We
extracted 90M token subsets for each language, shuffled them by
sentence and split them into training and validation sets (8-to-1
proportion). For LM training, we included the 50K most
frequent words in each corpus in the vocabulary, replacing the other
tokens with the 
UNK
symbol. The validation set
perplexity values we report below exclude unknown tokens.

\paragraph{RNN language models.}
We experimented with simple RNNs (sRNNs, \citealp{Elman:1990}), and
their most successful variant, long-short term memory models (LSTMs,
\citealp{Hochreiter:Schmidhuber:1997}). We use the PyTorch RNN
implementation.\footnote{\url{https://github.com/pytorch/examples/tree/master/word_language_model}} We trained the models with two hidden layer dimensionalities (650 and 200 units), and a range of batch sizes, learning rates and dropout rates. See SM for details on hyperparameter tuning. In general, a larger hidden layer size was the best predictor of lower 
perplexity. Given that our LSTMs outperformed our sRNNs, our 
discussion of the results will focus on the former; we will use the
terms LSTM and RNN interchangeably.\footnote{Detailed results for sRNNs can be found in the SM.} 

\paragraph{Baselines.}
We consider three baselines: first, a \textbf{unigram} baseline, which picks the most frequent form in the
training corpus out of the two candidate target forms (singular or
plural); second, a 5-gram model with Kneser-Ney smoothing (\textbf{KN},
\citealp{kneser1995improved}) trained using the IRSTLM package
\cite{federico2008irstlm} and queried using KenLM
\cite{heafield2011kenlm}; and third,  a \textbf{5-gram
  LSTM}, which only had access to windows of five tokens
\cite{chelba2017n}. Compared to KN, the 5-gram LSTM can generalize to
unseen n-grams thanks to its embedding layer and recurrent
connections. However, it cannot discover long-distance dependency
patterns that span more than five words. See SM for details on the hyperparameters of this baseline.

\paragraph{Human experiment in Italian.} We presented the full Italian
test set (119 original and 1071 nonce sentences) to human subjects
through the Amazon Mechanical Turk
interface.\footnote{\url{https://www.mturk.com/}} We picked Italian
because, being morphologically richer, it features more varied
long-distance constructions than English. Subjects were requested to
be native Italian speakers. They were presented with a sentence up
to and excluding the target. The singular and plural forms of the
target were presented below the sentence (in random order), and
subjects were asked to select the more plausible form.

To prevent long-distance agreement patterns from being too
salient, we mixed the test set with the same number of
filler sentences. We started from original fillers, which were random
treebank-extracted sentences up to a content word in singular
or plural form. We then generated nonce fillers from the original ones
using the procedure outlined in Section \ref{sec:long_distance}. A
control subset of 688 fillers was manually selected by a
linguistically-trained Italian native speaker as unambiguous cases. To make sure we were only using data from native (or at least highly proficient) Italian speakers, we filtered out the responses of subjects who chose the wrong target in more than 20\% of the fillers.

We collected on average 9.5 judgments for each item (minimum 5
judgments). To account for the variable number of judgments across sentences, accuracy
rates were first calculated within each sentence and then averaged
across sentences.

\section{Results}
\begin{table}[t]
  \centering
  \setlength{\tabcolsep}{5.1pt}
 
  \resizebox{0.96\linewidth}{!}{
  \begin{tabular}{lrrrr}
    \toprule
    & \multicolumn{1}{c}{\textsc{It}} & \multicolumn{1}{c}{\textsc{En}} & \multicolumn{1}{c}{\textsc{He}} & \multicolumn{1}{c}{\textsc{Ru}}\tabularnewline
 \midrule
\multicolumn{1}{l}{\#constructions} & 8 & 2 & 18 & 21\tabularnewline
    \multicolumn{1}{l}{\#original} & 119 & 41 & 373 & 442\tabularnewline
\midrule
\multicolumn{2}{l}{\textbf{Unigram}}\tabularnewline
Original & 54.6 & 65.9 & 67.8 & 60.2\tabularnewline
Nonce & 54.1 & 42.5 & 63.1 & 54.0\tabularnewline
  \midrule
  \multicolumn{2}{l}{\textbf{5-gram KN}}\tabularnewline
Original & 63.9    & 63.4     & 72.1  & 73.5  \tabularnewline
Nonce & 52.8   & 43.4    & 61.7    & 56.8   \tabularnewline
        \cmidrule{1-5}
         Perplexity & 147.8 & 168.9 & 122.0 & 166.6\tabularnewline
 \midrule
 \multicolumn{2}{l}{\textbf{5-gram LSTM}}\tabularnewline
Original & 81.8 & 70.2 & 90.9 & 91.5\tabularnewline
& $^{\pm3.2}$ & $^{\pm5.8}$ & $^{\pm1.2}$ & $^{\pm0.4}$\tabularnewline
Nonce & 78.0 & 58.2 & 77.5 & 85.7\tabularnewline
& $^{\pm1.3}$ & $^{\pm2.1}$ & $^{\pm0.8}$ & $^{\pm0.7}$\tabularnewline
 \cmidrule{1-5}
 Perplexity & 62.6 & 71.6 & 59.9 & 61.1\tabularnewline
 & $^{\pm0.2}$ & $^{\pm0.3}$ & $^{\pm0.2}$ & $^{\pm0.4}$\tabularnewline
 \midrule
 \multicolumn{2}{l}{\textbf{LSTM}}\tabularnewline
Original & 92.1 & 81.0 & 94.7 & 96.1\tabularnewline
 & $^{\pm1.6}$ & $^{\pm2.0}$ & $^{\pm0.4}$ & $^{\pm0.7}$\tabularnewline
 Nonce & 85.5 & 74.1 & 80.8 & 88.8\tabularnewline
    & $^{\pm0.7}$ & $^{\pm1.6}$ & $^{\pm0.8}$ & $^{\pm0.9}$\tabularnewline
\cmidrule{1-5}
    Perplexity & 45.2 & 52.1 & 42.5 & 48.9\tabularnewline
    & $^{\pm0.3}$ & $^{\pm0.3}$ & $^{\pm0.2}$ & $^{\pm0.6}$\tabularnewline
 \bottomrule
\end{tabular}
}
    \caption{Experimental results for all languages averaged across the five best models in terms of perplexity on the validation set. Original/Nonce rows report percentage accuracy, and the numbers in small print represent standard deviation within the five best models.}
\label{t:all_langs}
\end{table}

The overall results are reported in Table~\ref{t:all_langs}.
We report results averaged across the five models with the lowest validation perplexity, as well as standard deviations across these models. 
In summary, the LSTM clearly outperformed the other LMs. Rather surprisingly, its performance on nonce sentences was only moderately lower than on original ones; in Italian this gap was only 6.6\%.

The KN LM performed poorly; its accuracy on nonce sentences was comparable to that of the unigram baseline. This confirms that the number of the target in
nonce sentences cannot be captured by shallow n-gram patterns. The
5-gram LSTM model greatly improved over the KN baseline; its accuracy
dropped only modestly between the original and nonce sentences, demonstrating its syntactic generalization ability. Still,
the results are substantially below those of the LSTM with unlimited
history. This confirms that our test set contains hard long-distance
agreement dependencies, and, more importantly, that the more general
LSTM model can exploit broader contexts to learn about and
track long-distance syntactic relations. 

The increase in
accuracy scores across the three LMs (KN, 5-gram LSTM and unbounded-context LSTM) correlates
well with their validation perplexities in the language modeling task. We also found a strong correlation
between agreement accuracy and validation perplexity across all the
LSTM variants we explored in the hyperparameter search (68 models per
language), with Pearson correlation coefficients ranging from $r = -0.55$ in Hebrew to $r = -0.78$
in English ($p<0.001$ in all languages). This suggests that acquiring
abstract syntactic competence is a natural component of the skills
that improve the generic language modeling performance of RNNs.

\begin{table}
\centering
\begin{tabular}{llcc}
\toprule
 &  & N V V & V NP conj V\tabularnewline
 \midrule
Italian & Original & 93.3$_{\pm4.1}$  & 83.3$_{\pm10.4}$ \tabularnewline
 & Nonce & 92.5$_{\pm2.1}$ & 78.5$_{\pm1.7}$\tabularnewline
English & Original & 89.6$_{\pm3.6}$  & 67.5$_{\pm5.2}$ \tabularnewline
 & Nonce & 68.7$_{\pm0.9}$ &  82.5$_{\pm4.8}$\tabularnewline
Hebrew & Original & 86.7$_{\pm9.3}$  & 83.3$_{\pm5.9}$ \tabularnewline
 & Nonce & 65.7$_{\pm4.1}$ & 83.1$_{\pm2.8}$\tabularnewline
Russian & Original & - & 95.2$_{\pm1.9}$ \tabularnewline
 & Nonce & - & 86.7$_{\pm1.6}$\tabularnewline
 \bottomrule
\end{tabular}

\caption{LSTM accuracy in the constructions N~V~V (subject-verb agreement with an intervening embedded clause) and V~NP~conj~V (agreement between conjoined verbs separated by a complement of the first verb).}
\label{t:two_hard_constructions}
\end{table}

\paragraph{Differences across languages.} English was by far the hardest language. We conjecture that this is due to
its poorer morphology and higher POS ambiguity, which might not encourage
a generic language model to track abstract syntactic
configurations. There is an alternative hypothesis, however. We only extracted two constructions
for English, both of which can be argued to be linguistically complex: subject-verb
agreement with an intervening embedded clause, and agreement between two conjoined
verbs with a nominal complement intervening between the verbs. Yet the results on these two
constructions, comparable across languages (with the exception of
the subject-verb construction in Russian, which was not extracted), confirm that English is
particularly hard (Table~\ref{t:two_hard_constructions}). A qualitative inspection suggests that the low
accuracy in the verb conjunction case (67.5\%) is due to
ambiguous sentences such as ``if you \myuline{have} any questions or
\myuline{need}/\myuline{needs}'', where the target can be
re-interpreted as a noun that is acceptable in the relevant
context.\footnote{The nonce condition has higher accuracy because our substitution procedure in English tends to reduce POS ambiguity.}

In languages such as Italian and Russian, which have richer morphology and less ambiguity at the part-of-speech level than English, the LSTMs show much better accuracy and a
smaller gap between original and nonce sentences. These results are 
in line with human experimental studies that
found that richer morphology correlates with fewer agreement
attraction errors \citep{Lorimor2008}. The pattern of accuracy rates in general, and
the accuracy for the shared V NP conj V construction in particular, are consistent with the finding that Russian is less prone to human attraction errors
than Italian, which, in turn, shows less errors than English.

The largest drop in accuracy between original and nonce sentences occurred in Hebrew. A qualitative analysis of the data in this language suggests that this might be due to the numerical prevalence of a few constructions that can have multiple alternative readings, some of which can license the incorrect number. We leave a more systematic analysis of this finding for future research.

\begin{table*}
\centering
\begin{tabular}{lrrrrr}
\toprule
Construction & \#original & \multicolumn{2}{c}{Original} & \multicolumn{2}{c}{Nonce}\tabularnewline
 &  & Subjects & LSTM & Subjects & LSTM\tabularnewline
 \midrule
 \textsc{det} {[}AdjP{]} \textsc{noun} & 14 & 98.7 & 98.6$_{\pm3.2}$ & 98.1 & 91.7$_{\pm0.4}$\tabularnewline
 \textsc{noun} {[}RelC / PartP{]} clitic \textsc{verb} & 6 & 93.1 & 100$_{\pm0.0}$ & 95.4 & 97.8$_{\pm0.8}$\tabularnewline
 \textsc{noun} {[}RelC / PartP {]} \textsc{verb} & 27 & 97.0 & 93.3$_{\pm4.1}$ & 92.3 & 92.5$_{\pm2.1}$\tabularnewline
 \textsc{adj} {[}conjoined \textsc{adj}s{]} \textsc{adj} & 13 & 98.5 & 100$_{\pm0.0}$ & 98.0 & 98.1$_{\pm1.1}$\tabularnewline
 \textsc{noun} {[}AdjP{]} relpron \textsc{verb} & 10 & 95.9 & 98.0$_{\pm4.5}$ & 89.5 & 84.0$_{\pm3.3}$\tabularnewline
 \textsc{noun} {[}PP{]} \textsc{adverb} \textsc{adj} & 13 & 91.5 & 98.5$_{\pm3.4}$ & 79.4 & 76.9$_{\pm1.4}$\tabularnewline
 \textsc{noun} {[}PP{]} \textsc{verb} (participial) & 18 & 87.1 & 77.8$_{\pm3.9}$ & 73.4 & 71.1$_{\pm3.3}$\tabularnewline
 \textsc{verb} {[}NP{]} \textsc{conj} \textsc{verb} & 18 & 94.0 & 83.3$_{\pm10.4}$ & 86.8 & 78.5$_{\pm1.7}$\tabularnewline
\midrule
(Micro) average &  & 94.5 & 92.1$_{\pm1.6}$ & 88.4 & 85.5$_{\pm0.7}$\tabularnewline
\bottomrule
\end{tabular}
    \caption{Subject and LSTM accuracy on the Italian test set, by construction and averaged.}
\label{t:italian_mturk}
\end{table*}

\paragraph{Human results.} To put our results in context and provide a reasonable upper bound on
the LM performance, in particular for nonce sentences,
we next compare model performance to that of human subjects in Italian.

Table~\ref{t:italian_mturk} reports the accuracy of the LSTMs and the 
human subjects, grouped by construction.\footnote{The SM
  contains the results for the other languages broken down by
  construction. Note that Table~\ref{t:italian_mturk} reports
  linguistically intuitive construction labels. The corresponding POS
  patterns are (in same order as table rows): \textsc{det adj noun, noun verb
  pron verb, noun verb verb, adj adj cconj adj, noun adj punct pron
  verb, noun noun adv adj, noun noun verb, verb noun cconj verb.}}
There was a consistent gap in human accuracy between
original and nonce sentences (6.1\% on average). The gap in
accuracy between the human subjects and the model was quite small, and
was similar for original and nonce sentences (2.4\% and 2.9\%,
respectively).

In some of the harder constructions, particularly
subject-verb agreement with an embedded clause, the accuracy of the LSTMs on nonce
sentences was comparable to human accuracy (92.5$_{\pm2.1}$ vs.~92.3\%). To test whether the human subjects and the models struggle with the same sentences, we computed for each sentence (1) the number of times the human subjects selected the correct form of the target minus the number of times they selected the incorrect form, and (2) the difference in model log probability between the correct and incorrect form. The Spearman correlation between these quantities was significant, for both original ($p < 0.05$) and nonce sentences ($p < 0.001$). This indicates that humans were more likely to select the correct form in sentences in which the models were more confident in a correct prediction.

Moreover, some of the easiest and hardest constructions are the same for the human subjects and
the models. In the easy constructions \textsc{det}
{[}AdjP{]} \textsc{noun}\footnote{The relatively low nonce LSTM performance on this
construction is due to a few adjectives that could be re-interpreted as nouns.} and \textsc{adj} {[}conjoined \textsc{adj}s{]} \textsc{adj}, one or more adjectives that intervene between
the cue and the target agree in number with the target, providing 
shorter-distance evidence about its correct number. For example, in

\exg. un film \myuline{inutile} ma almeno \myuline{festivo} e \myuline{giovanile}\\
a movie useless but at.least festive and youthful\\
``A useless but at least festive and youthful movie''

\noindent{}the adjective ``festivo'' is marked for singular number,
 offering a nearer reference for the target number than the cue
 ``inutile''. At the other end, \textsc{noun} {[}PP{]} \textsc{verb} (participial)
 and \textsc{noun} {[}PP{]} \textsc{adverb adj} are difficult. Particularly in the
nonce condition, where semantics is unhelpful or even misleading, the
target could easily be interpreted as a modifier of the noun embedded
in the preceding prepositional phrase. For example, for the nonce
case:

\exg. \myuline{orto} di \myuline{regolamenti} davvero \myuline{pedonale/i}\\
orchard of rules truly pedestrian\\
``truly pedestrian orchard of rules''

\noindent{}both the subjects and the model preferred to treat ``pedestrian'' as
a modifier of ``rules'' (``orchard of truly pedestrian rules''),
resulting in the wrong agreement given the intended syntactic
structure. 

\begin{figure}[tb]
  \centering 
  \includegraphics[width=20.1em]{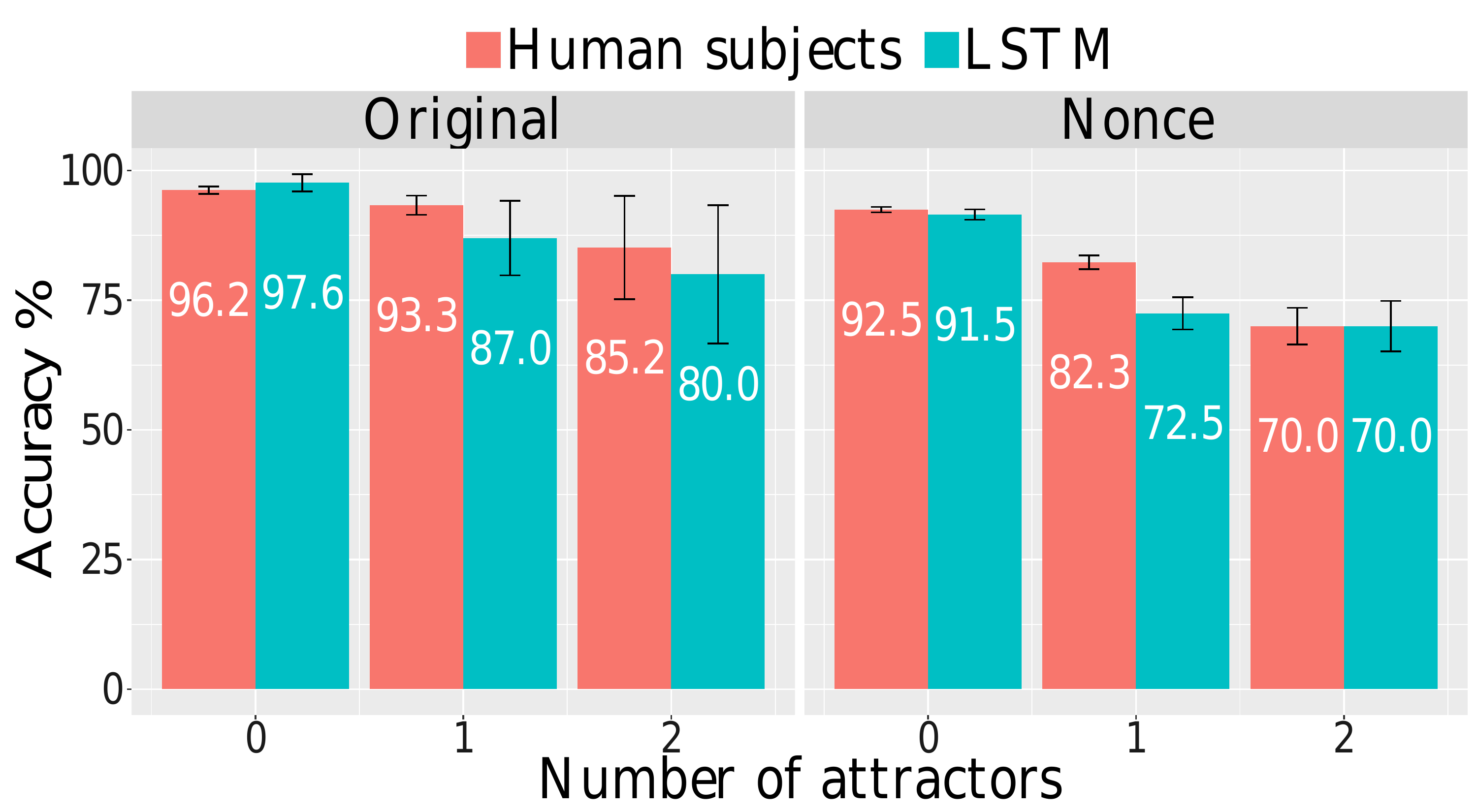}
  \caption{Accuracy by number of attractors in Italian. Human
    performance is shown in red and LSTM in blue (median model among top 5 ranked
    by perplexity). Error bars show standard error.}
  \label{fig:attractors}
\end{figure}

\paragraph{Attractors.} 
We define attractors as words with the same POS as the cue but
the opposite number, which intervene in the linear order of the sentence
between the cue and the target. Attractors constitute an obvious
challenge for agreement processing \citep{Bock:Miller:1991}. We
show how their presence affects human and model behavior in
Fig.~\ref{fig:attractors}. We limit our analysis to a maximum of two
attractors, since there were only two original sentences in the test
corpus with three attractors or more. Both model and human accuracies degraded with
the number of attractors; the drop in accuracy was sharper in the nonce condition. While the
model performed somewhat worse than humans, the overall pattern was comparable.

Our results suggest that the LSTM is  quite robust to the
presence of  attractors, in contrast to what was reported by
\citet{Linzen:etal:2016}. We directly compared our English LSTM LM to theirs
by predicting verb number on the \citet{Linzen:etal:2016} test set. We
extracted sentences where all of the words between subject and verb were in
our LM vocabulary. Out of those sentences, we sampled 2000 sentences with 0, 1 and 2 attractors and kept all the sentences with 3 and 4 attractors
(1329 and 347 sentences, respectively). To ensure that our training
set and Linzen's test set do not overlap (both are based on Wikipedia
texts), we filtered out all of test sentences that appeared in our training
data (187 sentences).  

\begin{figure}[tb]
  \centering 
  \includegraphics[width=20.1em]{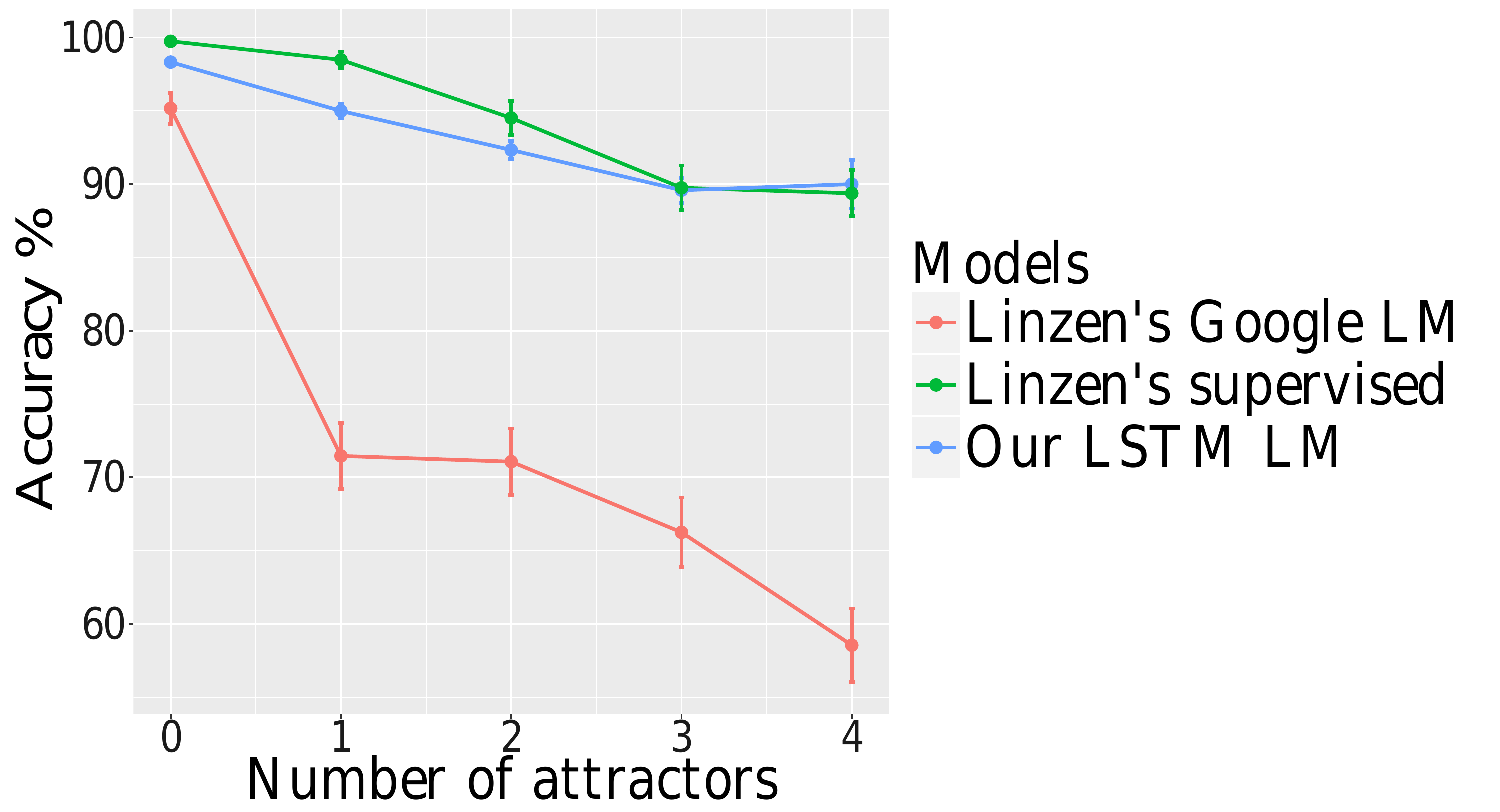}
  \caption{Linzen's attractor set. Our LM-trained LSTM (blue; ``median'' model) compared to their LSTM with explicit number supervision (green) and their best LM-trained LSTM (red).}
  \label{fig:attractors_tal}
\end{figure}

Fig.~\ref{fig:attractors_tal} compares our results 
to the results of the best LM-trained model in \citet{Linzen:etal:2016}
(their ``Google LM'').\footnote{These subject-verb agreement results
  are in general higher than for our own subject-verb agreement
  construction (\textsc{noun verb verb}) because the latter always includes an
  embedded clause, and it is therefore harder on average.} Not only
did our LM greatly outperform theirs, but it approached the
performance of their supervised model.\footnote{Similarly high performance of LM-trained RNNs on Linzen's dataset was recently reported by \newcite{yogatama2018memory}.} This difference in results
points to the importance of careful tuning of LM-trained LSTMs, although
we must leave to a further study a more detailed understanding of
which differences crucially determine our better performance.

\section{Related work}

Early work showed that RNNs can, to a certain degree, handle data
generated by context-free and even context-sensitive grammars
\citep[e.g.,][]{Elman:1991,Elman:1993,Rohde:Plaut:1997,Christiansen:Chater:1999,Gers:Schmidhuber:2001,Cartling:2008}. These
experiments were based on small and controlled artificial
languages, in which complex hierarchical phenomena were often
over-represented compared to natural languages.

Our work, which is based on naturally occurring data, is most closely related to that of \newcite{Linzen:etal:2016}
and \citet{Bernardy:Lappin:2017}, which we discussed in the
introduction.  Other recent work has focused on the morphological and
grammatical knowledge that RNN-based machine-translation systems and
sentence embeddings encode, typically by training classifiers to
decode various linguistic properties from hidden states of the network
\citep[e.g.,][]{Adi:etal:2017,Belinkov:etal:2017,Shi:etal:2016}, or
looking at whether the end-to-end system correctly translates
sentences with challenging constructions \citep{Sennrich:2017}.

Previous work in neurolinguistics and psycholinguistics used jabberwocky, or pseudo-word, sentences to probe how speakers process syntactic information \citep{friederici2000,moro2001syntax,johnson2013}. Such sentences are obtained by substituting original words with morphologically and phonologically acceptable nonce forms. We are not aware of work that used nonce sentences made of real words to evaluate the syntactic abilities of models or human subjects. As a proof of concept, \newcite{pereira2000} and, later, \newcite{Mikolov:2012} computed the probability of  Chomsky's famous ``colorless green ideas'' sentence using a class-based bigram LM and an RNN, respectively, and showed that it is much higher than the probability of its shuffled ungrammatical variants.

\section{Conclusion}

We ran an extensive analysis of the abilities of RNNs trained on a
generic language-modeling task to predict long-distance number
agreement. Results were consistent across four languages and a number
of constructions. They were above strong baselines even in the
challenging case of nonsense sentences, and not far from human
performance. We are not aware of other collections of human
long-distance agreement judgments on nonsensical sentences, and we
thus consider our publicly available data set an important
contribution of our work, of interest to students of human language processing in general.

The constructions we considered are quite infrequent (according to a
rough estimate based on the treebanks, the language in which they are most common is
Hebrew, and even there they occur with average 0.8\% sentence
frequency). Moreover, they vary in the contexts that separate the cue and
the target. So, RNNs are not simply memorizing frequent morphosyntactic
sequences (which would already be impressive, for systems learning
from raw text). We tentatively conclude that LM-trained RNNs can
construct abstract grammatical representations of their input. This,
in turn, suggests that the input itself contains enough information to
trigger some form of syntactic learning in a system, such as an RNN,
that does not contain an explicit prior bias in favour of syntactic
structures.

In future work, we would like to better understand what kind of
syntactic information RNNs are encoding, and how. On the one hand, we
plan to adapt methods to inspect information flow across RNN states
\citep[e.g.,][]{Hupkes:etal:2017}. On the other, we would like to
expand our empirical investigation by focusing on other long-distance
phenomena, such as overt case assignment \citep{Blake:2001} or
parasitic gap licensing \citep{Culicover:Postal:2001}. While it is more
challenging to extract reliable examples of such phenomena from
corpora, their study would probe more sophisticated syntactic
capabilities, possibly even shedding light on the theoretical analysis
of the underlying linguistic structures. Finally, it may be useful
to complement the corpus-driven approach used in the current paper with constructed evaluation sentences that isolate particular syntactic phenomena, independent of their frequency in a natural corpus, as is common in psycholinguistics \cite{enguehard2017exploring}.

\section*{Acknowledgements}

We thank the reviewers, Germ\'{a}n Kruszewski, Gerhard J\"ager, Adam Li\v{s}ka, Tomas Mikolov, Gemma Boleda, Brian Dillon, Cristophe Pallier, Roberto Zamparelli and the Paris Syntax and Semantics Colloquium audience for feedback and advice.

\bibliographystyle{acl_natbib}
\bibliography{marco,kristina}

\begin{thebibliography}{}
\expandafter\ifx\csname natexlab\endcsname\relax\def\natexlab#1{#1}\fi

\bibitem[{Adi et~al.(2017)Adi, Kermany, Belinkov, Lavi, and
  Goldberg}]{Adi:etal:2017}
Yossi Adi, Einat Kermany, Yonatan Belinkov, Ofer Lavi, and Yoav Goldberg. 2017.
\newblock Fine-grained analysis of sentence embeddings using auxiliary
  prediction tasks.
\newblock In {\em Proceedings of ICLR Conference Track\/}. Toulon, France.
\newblock Published online:
  \url{https://openreview.net/group?id=ICLR.cc/2017/conference}.

\bibitem[{Belinkov et~al.(2017)Belinkov, Durrani, Dalvi, Sajjad, and
  Glass}]{Belinkov:etal:2017}
Yonatan Belinkov, Nadir Durrani, Fahim Dalvi, Hassan Sajjad, and James Glass.
  2017.
\newblock What do neural machine translation models learn about morphology?
\newblock In {\em Proceedings of ACL\/}. Vancouver, Canada, pages 861--872.

\bibitem[{Bernardy and Lappin(2017)}]{Bernardy:Lappin:2017}
{Jean-Philippe} Bernardy and Shalom Lappin. 2017.
\newblock Using deep neural networks to learn syntactic agreement.
\newblock {\em Linguistic Issues in Language Technology\/} 15(2):1--15.

\bibitem[{Blake(2001)}]{Blake:2001}
Barry Blake. 2001.
\newblock {\em Case\/}.
\newblock MIT Press, Cambridge, MA.

\bibitem[{Bock and Miller(1991)}]{Bock:Miller:1991}
Kathryn Bock and Carol Miller. 1991.
\newblock Broken agreement.
\newblock {\em Cognitive Psychology\/} 23(1):45--93.

\bibitem[{Cartling(2008)}]{Cartling:2008}
Bo~Cartling. 2008.
\newblock On the implicit acquisition of a context-free grammar by a simple
  recurrent neural network.
\newblock {\em Neurocomputing\/} 71:1527--1537.

\bibitem[{Chelba et~al.(2017)Chelba, Norouzi, and Bengio}]{chelba2017n}
Ciprian Chelba, Mohammad Norouzi, and Samy Bengio. 2017.
\newblock N-gram language modeling using recurrent neural network estimation.
\newblock {\em arXiv preprint arXiv:1703.10724\/} .

\bibitem[{Chomsky(1957)}]{Chomsky:1957}
Noam Chomsky. 1957.
\newblock {\em Syntactic Structures\/}.
\newblock Mouton, Berlin, Germany.

\bibitem[{Christiansen and Chater(1999)}]{Christiansen:Chater:1999}
Morten Christiansen and Nick Chater. 1999.
\newblock Toward a connectionist model of recursion in human linguistic
  performance.
\newblock {\em Cognitive Science\/} 23(2):157--205.

\bibitem[{Cross and Huang(2016)}]{Cross:Huang:2016}
James Cross and Liang Huang. 2016.
\newblock Incremental parsing with minimal features using bi-directional
  {LSTM}.
\newblock In {\em Proceedings of ACL (Short Papers)\/}. Berlin, Germany, pages
  32--37.

\bibitem[{Culicover and Postal(2001)}]{Culicover:Postal:2001}
Peter Culicover and Paul Postal, editors. 2001.
\newblock {\em Parasitic gaps\/}.
\newblock MIT Press, Cambridge, MA.

\bibitem[{Elman(1990)}]{Elman:1990}
Jeffrey Elman. 1990.
\newblock Finding structure in time.
\newblock {\em Cognitive Science\/} 14:179--211.

\bibitem[{Elman(1991)}]{Elman:1991}
Jeffrey Elman. 1991.
\newblock Distributed representations, simple recurrent networks, and
  grammatical structure.
\newblock {\em Machine Learning\/} 7:195--225.

\bibitem[{Elman(1993)}]{Elman:1993}
Jeffrey Elman. 1993.
\newblock Learning and development in neural networks: The importance of
  starting small.
\newblock {\em Cognition\/} 48:71--99.

\bibitem[{Enguehard et~al.(2017)Enguehard, Goldberg, and
  Linzen}]{enguehard2017exploring}
\'{E}mile Enguehard, Yoav Goldberg, and Tal Linzen. 2017.
\newblock Exploring the syntactic abilities of {RNNs} with multi-task learning.
\newblock In {\em {Proceedings of the 21st Conference on Computational Natural
  Language Learning (CoNLL 2017)}\/}. pages 3--14.

\bibitem[{Everaert et~al.(2015)Everaert, Huybregts, Chomsky, Berwick, and
  Bolhuis}]{Everaert:etal:2015}
Martin Everaert, Marinus Huybregts, Noam Chomsky, Robert Berwick, and Johan
  Bolhuis. 2015.
\newblock Structures, not strings: Linguistics as part of the cognitive
  sciences.
\newblock {\em Trends in Cognitive Sciences\/} 19(12):729--743.

\bibitem[{Federico et~al.(2008)Federico, Bertoldi, and
  Cettolo}]{federico2008irstlm}
Marcello Federico, Nicola Bertoldi, and Mauro Cettolo. 2008.
\newblock Irstlm: An open source toolkit for handling large scale language
  models.
\newblock In {\em Ninth Annual Conference of the International Speech
  Communication Association\/}. pages 1618--1621.

\bibitem[{Friederici et~al.(2000)Friederici, Meyer, and {Von
  Cramon}}]{friederici2000}
Angela~D. Friederici, Martin Meyer, and D.~Yves {Von Cramon}. 2000.
\newblock {Auditory language comprehension: An event-related fMRI study on the
  processing of syntactic and lexical information}.
\newblock {\em Brain and Language\/} 74(2):289--300.

\bibitem[{Gers and Schmidhuber(2001)}]{Gers:Schmidhuber:2001}
Felix Gers and J\"{u}rgen Schmidhuber. 2001.
\newblock {LSTM} recurrent networks learn simple context-free and
  context-sensitive languages.
\newblock {\em IEEE Transactions on Neural Networks\/} 12(6):1333--1340.

\bibitem[{Graves(2012)}]{Graves:2012}
Alex Graves. 2012.
\newblock {\em Supervised Sequence Labelling with Recurrent Neural Networks\/}.
\newblock Springer, Berlin.

\bibitem[{Heafield(2011)}]{heafield2011kenlm}
Kenneth Heafield. 2011.
\newblock Kenlm: Faster and smaller language model queries.
\newblock In {\em Proceedings of the Sixth Workshop on Statistical Machine
  Translation\/}. Association for Computational Linguistics, pages 187--197.

\bibitem[{Hochreiter and Schmidhuber(1997)}]{Hochreiter:Schmidhuber:1997}
Sepp Hochreiter and J\"{u}rgen Schmidhuber. 1997.
\newblock Long short-term memory.
\newblock {\em Neural Computation\/} 9(8):1735--178--.

\bibitem[{{Hupkes} et~al.(2017){Hupkes}, {Veldhoen}, and
  {Zuidema}}]{Hupkes:etal:2017}
Dieuwke {Hupkes}, Sara {Veldhoen}, and Willem {Zuidema}. 2017.
\newblock Visualisation and diagnostic classifiers reveal how recurrent and
  recursive neural networks process hierarchical structure.
\newblock \url{http://arxiv.org/abs/1711.10203}.

\bibitem[{Johnson and Goldberg(2013)}]{johnson2013}
Matt~A. Johnson and Adele~E. Goldberg. 2013.
\newblock Evidence for automatic accessing of constructional meaning:
  Jabberwocky sentences prime associated verbs.
\newblock {\em Language and Cognitive Processes\/} 28(10):1439--1452.

\bibitem[{Kiperwasser and Goldberg(2016)}]{Kiperwasser:Goldberg:2016}
Eliyahu Kiperwasser and Yoav Goldberg. 2016.
\newblock Simple and accurate dependency parsing using bidirectional {LSTM}
  feature representations.
\newblock {\em Transactions of the Association for Computational Linguistics\/}
  4:313--327.

\bibitem[{Kneser and Ney(1995)}]{kneser1995improved}
Reinhard Kneser and Hermann Ney. 1995.
\newblock Improved backing-off for m-gram language modeling.
\newblock In {\em Acoustics, Speech, and Signal Processing, 1995. ICASSP-95.,
  1995 International Conference on\/}. IEEE, volume~1, pages 181--184.

\bibitem[{Linzen et~al.(2016)Linzen, Dupoux, and Goldberg}]{Linzen:etal:2016}
Tal Linzen, Emmanuel Dupoux, and Yoav Goldberg. 2016.
\newblock Assessing the ability of {LSTM}s to learn syntax-sensitive
  dependencies.
\newblock {\em Transactions of the Association for Computational Linguistics\/}
  4:521--535.

\bibitem[{Lorimor et~al.(2008)Lorimor, Bock, Zalkind, Sheyman, and
  Beard}]{Lorimor2008}
Heidi Lorimor, Kathryn Bock, Ekaterina Zalkind, Alina Sheyman, and Robert
  Beard. 2008.
\newblock {Agreement and attraction in Russian}.
\newblock {\em Language and Cognitive Processes\/} 23(6):769--799.

\bibitem[{Mikolov(2012)}]{Mikolov:2012}
Tomas Mikolov. 2012.
\newblock {\em Statistical language models based on neural networks\/}.
\newblock Dissertation, Brno University of Technology.

\bibitem[{Mikolov et~al.(2010)Mikolov, Karafi{\'{a}}t, Burget, Cernock{\'{y}},
  and Khudanpur}]{Mikolov:etal:2010}
Tomas Mikolov, Martin Karafi{\'{a}}t, Luk{\'{a}}s Burget, Jan Cernock{\'{y}},
  and Sanjeev Khudanpur. 2010.
\newblock Recurrent neural network based language model.
\newblock In {\em Proceedings of INTERSPEECH\/}. Makuhari, Japan, pages
  1045--1048.

\bibitem[{Mikolov et~al.(2013)Mikolov, Yih, and Zweig}]{Mikolov:etal:2013a}
Tomas Mikolov, Wen-tau Yih, and Geoffrey Zweig. 2013.
\newblock Linguistic regularities in continuous space word representations.
\newblock In {\em Proceedings of NAACL\/}. Atlanta, Georgia, pages 746--751.

\bibitem[{Moro et~al.(2001)Moro, Tettamanti, Perani, Donati, Cappa, and
  Fazio}]{moro2001syntax}
Andrea Moro, Marco Tettamanti, Daniela Perani, Caterina Donati, Stefano~F
  Cappa, and Ferruccio Fazio. 2001.
\newblock Syntax and the brain: disentangling grammar by selective anomalies.
\newblock {\em Neuroimage\/} 13(1):110--118.

\bibitem[{Nivre et~al.(2016)Nivre, de~Marneffe, Ginter, Goldberg, Hajic,
  Manning, McDonald, Petrov, Pyysalo, Silveira, Tsarfaty, and
  Zeman}]{Nivre:2016}
Joakim Nivre, Marie-Catherine de~Marneffe, Filip Ginter, Yoav Goldberg, Jan
  Hajic, Christopher~D. Manning, Ryan McDonald, Slav Petrov, Sampo Pyysalo,
  Natalia Silveira, Reut Tsarfaty, and Daniel Zeman. 2016.
\newblock Universal dependencies v1: A multilingual treebank collection.
\newblock In Nicoletta Calzolari~(Conference Chair), Khalid Choukri, Thierry
  Declerck, Sara Goggi, Marko Grobelnik, Bente Maegaard, Joseph Mariani, Helene
  Mazo, Asuncion Moreno, Jan Odijk, and Stelios Piperidis, editors, {\em
  Proceedings of the Tenth International Conference on Language Resources and
  Evaluation (LREC 2016)\/}. European Language Resources Association (ELRA),
  Paris, France.

\bibitem[{Pereira(2000)}]{pereira2000}
Fernando Pereira. 2000.
\newblock Formal grammar and information theory: together again?
\newblock {\em Philosophical Transactions of the Royal Society of London A:
  Mathematical, Physical and Engineering Sciences\/} 358(1769):1239--1253.

\bibitem[{Rohde and Plaut(1997)}]{Rohde:Plaut:1997}
Douglas Rohde and David Plaut. 1997.
\newblock Simple recurrent networks and natural language: How important is
  starting small?
\newblock In {\em Proceedings of CogSci\/}. Stanford, CA, pages 656--661.

\bibitem[{Schmid(1995)}]{Schmid:1995}
Helmut Schmid. 1995.
\newblock Improvements in part-of-speech tagging with an application to
  {G}erman.
\newblock In {\em Proceedings of the EACL-SIGDAT Workshop\/}. Dublin, Ireland.

\bibitem[{Sennrich(2017)}]{Sennrich:2017}
Rico Sennrich. 2017.
\newblock How grammatical is character-level neural machine translation?
  assessing {MT} quality with contrastive translation pairs.
\newblock In {\em Proceedings of EACL (Short Papers)\/}. Valencia, Spain, pages
  376--382.

\bibitem[{Shi et~al.(2016)Shi, Padhi, and Knight}]{Shi:etal:2016}
Xing Shi, Inkit Padhi, and Kevin Knight. 2016.
\newblock Does string-based neural {MT} learn source syntax?
\newblock In {\em Proceedings of EMNLP\/}. Austin, Texas, pages 1526--1534.

\bibitem[{Wu et~al.(2016)Wu, Schuster, Chen, Le, Norouzi, Macherey, Krikun,
  Cao, Gao, Macherey, Klingner, Shah, Johnson, Liu, Kaiser, Gouws, Kato, Kudo,
  Kazawa, Stevens, Kurian, Patil, Wang, Young, Smith, Riesa, Rudnick, Vinyals,
  Corrado, Hughes, and Dean}]{Wu:etal:2016}
Yonghui Wu, Mike Schuster, Zhifeng Chen, Quoc Le, Mohammad Norouzi, Wolfgang
  Macherey, Maxim Krikun, Yuan Cao, Qin Gao, Klaus Macherey, Jeff Klingner,
  Apurva Shah, Melvin Johnson, Xiaobing Liu, Lukasz Kaiser, Stephan Gouws,
  Yoshikiyo Kato, Taku Kudo, Hideto Kazawa, Keith Stevens, George Kurian,
  Nishant Patil, Wei Wang, Cliff Young, Jason Smith, Jason Riesa, Alex Rudnick,
  Oriol Vinyals, Greg Corrado, Macduff Hughes, and Jeffrey Dean. 2016.
\newblock Google's neural machine translation system: Bridging the gap between
  human and machine translation.
\newblock \url{http://arxiv.org/abs/1609.08144}.

\bibitem[{Yogatama et~al.(2018)Yogatama, Miao, Melis, Ling, Kuncoro, Dyer, and
  Blunsom}]{yogatama2018memory}
Dani Yogatama, Yishu Miao, Gabor Melis, Wang Ling, Adhiguna Kuncoro, Chris
  Dyer, and Phil Blunsom. 2018.
\newblock \href{https://openreview.net/forum?id=SkFqf0lAZ}{Memory architectures
  in recurrent neural network language models}.
\newblock In {\em International Conference on Learning Representations\/}.
\newblock \url{https://openreview.net/forum?id=SkFqf0lAZ}.

\bibitem[{Zhang et~al.(2017)Zhang, Cheng, and Lapata}]{Zhang:etal:2017}
Xingxing Zhang, Jianpeng Cheng, and Mirella Lapata. 2017.
\newblock Dependency parsing as head selection.
\newblock In {\em Proceedings of EACL\/}. Valencia, Spain, pages 665--676.

\end{thebibliography}

\end{document}

\maketitle

\section*{Treebank data post-processing}

In Hebrew, for consistency with the Wikipedia training data, we
tokenized the sentences based on UD morpheme annotation, by
treating orthographic clitics that can precede the word (conjunctions,
prepositions and definite articles) as separate tokens.  We also
modified the morphological annotation to reflect the definiteness of
nouns with possessive suffixes, to ensure that indefinite nouns are
only substituted by other indefinites. Finally, since the Hebrew treebank is the smallest one in our set, we use the automatically parsed Hebrew Wikipedia to extract words annotated with their morphological information. 

For English, the UD treebank only marks third-person singular present
tense verbs. We added explicit plural features to the other finite
present tense verb forms.

\section*{Hyperparameters and further model details}

\paragraph{LSTM} For LSTMs we explored the following hyperparameters,
for a total of 68 combinations (we only considered two-layer models):
\begin{enumerate}
\item hidden and embedding size: 200 and 650
\item batch size: 20 (only for 200-units models), 64, 128
\item dropout rate: 0.0, 0.1, 0.2, 0.4 (for 650-units models only)
\item learning rate: 1.0, 5.0, 10.0, 20.0
\end{enumerate}

\setlength{\tabcolsep}{3.8pt}
\begin{table}[p]
  \begin{tabular}{ccccccccc}
\toprule
 & \multicolumn{3}{c}{accuracy} & ppl & hidden/ & batch size & dropout rate & learning rate\tabularnewline
 & total & orig & nonce &  & embedding size &  &  & \tabularnewline
 \midrule
Italian & 85.4 & 90.8 & 84.8 & 44.9 & 650 & 64 & 0.2 & 20.0\tabularnewline
 & 87.2 & 93.3 & 86.6 & 45.0 & 650 & 64 & 0.2 & 10.0\tabularnewline
 & 86.2 & 93.3 & 85.4 & 45.1 & 650 & 128 & 0.2 & 20.0\tabularnewline
 & 86.5 & 93.3 & 85.7 & 45.3 & 650 & 64 & 0.1 & 20.0\tabularnewline
 & 85.6 & 89.9 & 85.2 & 45.6 & 650 & 128 & 0.1 & 20.0\tabularnewline
 \midrule
English & 72.7 & 80.5 & 71.8 & 51.9 & 650 & 128 & 0.2 & 20.0\tabularnewline
 & 76.8 & 82.9 & 76.2 & 51.9 & 650 & 64 & 0.2 & 10.0\tabularnewline
 & 75.4 & 78.0 & 75.1 & 52.1 & 650 & 64 & 0.2 & 20.0\tabularnewline
 & 74.1 & 80.5 & 73.4 & 52.1 & 650 & 64 & 0.1 & 20.0\tabularnewline
 & 74.9 & 82.9 & 74.0 & 52.6 & 650 & 128 & 0.1 & 20.0\tabularnewline
 \midrule
Hebrew & 81.3 & 95.2 & 79.8 & 42.3 & 650 & 64 & 0.1 & 20.0\tabularnewline
 & 82.3 & 94.4 & 81.0 & 42.4 & 650 & 128 & 0.2 & 20.0\tabularnewline
 & 83.2 & 94.9 & 81.9 & 42.5 & 650 & 64 & 0.2 & 20.0\tabularnewline
 & 82.3 & 94.9 & 80.8 & 42.7 & 650 & 64 & 0.1 & 10.0\tabularnewline
 & 81.7 & 94.4 & 80.3 & 42.7 & 650 & 128 & 0.1 & 20.0\tabularnewline
 \midrule
Russian & 90.8 & 96.6 & 90.2 & 48.1 & 650 & 64 & 0.2 & 20.0\tabularnewline
 & 89.6 & 97.1 & 88.8 & 48.6 & 650 & 64 & 0.1 & 20.0\tabularnewline
 & 88.6 & 95.9 & 87.8 & 49.1 & 650 & 64 & 0.1 & 10.0\tabularnewline
 & 89.0 & 95.2 & 88.3 & 49.2 & 650 & 128 & 0.1 & 20.0\tabularnewline
 & 89.8 & 95.7 & 89.1 & 49.6 & 650 & 128 & 0.2 & 20.0\tabularnewline
 \bottomrule
  \end{tabular}
\caption{Top 5 LSTMs.}
\label{t:lstm}
\end{table}

The LSTM results reported in the paper are averaged over the 5 models
that achieved the lowest validation perplexity after 40 training epochs. Their individual accuracies are reported in Table \ref{t:lstm}.

\paragraph{sRNN} For sRNNs we explored the following hyperparameters, for a total of 24 models:

\begin{enumerate}
\item hidden/embedding size: 650
\item batch size: 64, 128
\item dropout rate: 0.0, 0.1, 0.2
\item learning rate: 0.5, 1.0, 5.0, 10.0
\end{enumerate}

Results obtained in all languages by the top-5 sRNNs by perplexity after 60 training epochs are reported in Table \ref{t:srnn}.

\begin{table}[p]
\begin{tabular}{ccccccccc}
\toprule
 & \multicolumn{3}{c}{accuracy} & ppl & hidden/ & batch size & dropout rate & learning rate\tabularnewline
 & total & orig & nonce &  & embedding size &  &  & \tabularnewline
  \midrule
Italian & 77.2 & 86.6 & 76.2 & 64.0 & 650 & 64 & 0.0 & 1.0\tabularnewline
 & 77.7 & 84.0 & 77.0 & 64.5 & 650 & 128 & 0.0 & 1.0\tabularnewline
 & 76.8 & 84.0 & 76.0 & 65.8 & 650 & 64 & 0.0 & 0.5\tabularnewline
 & 78.5 & 85.7 & 77.7 & 66.1 & 650 & 64 & 0.1 & 1.0\tabularnewline
 & 78.2 & 87.4 & 77.1 & 67.8 & 650 & 64 & 0.1 & 0.5\tabularnewline
 \midrule
English & 65.1 & 73.2 & 64.2 & 69.8 & 650 & 64 & 0.0 & 1.0\tabularnewline
 & 65.6 & 82.9 & 63.7 & 70.6 & 650 & 128 & 0.0 & 1.0\tabularnewline
 & 69.5 & 70.7 & 69.4 & 72.9 & 650 & 64 & 0.1 & 1.0\tabularnewline
 & 63.9 & 68.3 & 63.4 & 73.3 & 650 & 64 & 0.0 & 0.5\tabularnewline
 & 65.9 & 75.6 & 64.8 & 74.4 & 650 & 128 & 0.1 & 1.0\tabularnewline
 \midrule
Hebrew & 75.8 & 88.5 & 74.4 & 60.9 & 650 & 128 & 0.0 & 1.0\tabularnewline
 & 75.5 & 88.7 & 74.0 & 62.8 & 650 & 64 & 0.0 & 1.0\tabularnewline
 & 74.8 & 88.2 & 73.3 & 63.3 & 650 & 64 & 0.0 & 0.5\tabularnewline
 & 75.7 & 88.2 & 74.3 & 63.7 & 650 & 64 & 0.1 & 1.0\tabularnewline
 & 76.0 & 87.9 & 74.6 & 63.9 & 650 & 128 & 0.1 & 1.0\tabularnewline
 \midrule
Russian & 82.4 & 87.6 & 81.9 & 65.1 & 650 & 64 & 0.1 & 1.0\tabularnewline
 & 81.6 & 87.1 & 80.9 & 65.6 & 650 & 64 & 0.0 & 1.0\tabularnewline
 & 81.3 & 86.4 & 80.7 & 68.2 & 650 & 128 & 0.0 & 5.0\tabularnewline
 & 81.7 & 87.1 & 81.1 & 68.4 & 650 & 128 & 0.1 & 10.0\tabularnewline
 & 82.0 & 87.6 & 81.3 & 68.6 & 650 & 128 & 0.1 & 1.0\tabularnewline
 \bottomrule
\end{tabular}
\caption{Top 5 sRNNs.}
\label{t:srnn}
\end{table}

\paragraph{5-gram LSTM} The 5-gram LSTM is trained as a count-based n-gram model, that is,
during training it observes all 5-grams in the text, moving ahead of
one token at each time step. For each 5-gram, the LSTM is
re-initialized with a blank (all zeroes) hidden state.  We tested only
models with 650 units in the hidden and embedding layers, and explored
the following hyperparameters, for a total of 24 models:
\begin{enumerate}
\item batch size: 256, 512, 1024
\item dropout rate: 0.0, 0.1, 0.2
\item learning rate: 1.0, 5.0, 10.0, 20.0
\end{enumerate}

The 5-gram LSTM results reported in the paper are averaged over the 5 models
that achieved the lowest validation perplexities, reported in Table
\ref{t:ngram_lstm}.

\begin{table}[t]
\centering
\begin{tabular}{ccccccccc}
\toprule
 & \multicolumn{3}{c}{accuracy} & ppl & hidden/ & batch size & dropout rate & learning rate\tabularnewline
 & total & orig & nonce &  & embedding size &  &  & \tabularnewline
 \midrule
Italian & 79.5 & 83.2 & 79.1 & 62.4 & 650 & 1024 & 0.2 & 10.0\tabularnewline
 & 78.5 & 77.3 & 78.6 & 62.4 & 650 & 256 & 0.2 & 5.0\tabularnewline
 & 76.6 & 79.8 & 76.3 & 62.4 & 650 & 512 & 0.2 & 5.0\tabularnewline
 & 77.8 & 84.0 & 77.1 & 62.7 & 650 & 1024 & 0.2 & 5.0\tabularnewline
 & 79.7 & 84.9 & 79.1 & 62.9 & 650 & 1024 & 0.1 & 10.0\tabularnewline
 \midrule
English & 58.0 & 70.7 & 56.6 & 71.6 & 650 & 512 & 0.2 & 5.0\tabularnewline
 & 61.5 & 65.9 & 61.0 & 71.7 & 650 & 1024 & 0.1 & 10.0\tabularnewline
 & 56.6 & 63.4 & 55.8 & 71.8 & 650 & 1024 & 0.2 & 10.0\tabularnewline
 & 61.5 & 78.0 & 59.6 & 71.8 & 650 & 1024 & 0.2 & 5.0\tabularnewline
 & 59.5 & 73.2 & 58.0 & 72.5 & 650 & 1024 & 0.1 & 5.0\tabularnewline
 \midrule
Hebrew & 78.4 & 90.1 & 77.1 & 59.7 & 650 & 256 & 0.1 & 5.0\tabularnewline
 & 80.1 & 92.5 & 78.7 & 59.7 & 650 & 1024 & 0.2 & 10.0\tabularnewline
 & 79.2 & 90.3 & 77.9 & 59.8 & 650 & 256 & 0.2 & 5.0\tabularnewline
 & 77.8 & 89.8 & 76.5 & 60.1 & 650 & 512 & 0.1 & 5.0\tabularnewline
 & 78.7 & 91.7 & 77.2 & 60.1 & 650 & 1024 & 0.1 & 10.0\tabularnewline
 \midrule
Russian & 86.2 & 91.9 & 85.5 & 61.1 & 650 & 1024 & 0.2 & 10.0\tabularnewline
 & 86.7 & 91.0 & 86.3 & 61.6 & 650 & 512 & 0.1 & 10.0\tabularnewline
 & 87.0 & 91.6 & 86.5 & 61.7 & 650 & 512 & 0.2 & 5.0\tabularnewline
 & 85.6 & 91.4 & 85.0 & 62.5 & 650 & 1024 & 0.2 & 5.0\tabularnewline
 & 85.9 & 91.9 & 85.2 & 62.8 & 650 & 512 & 0.2 & 10.0\tabularnewline
 \bottomrule
\end{tabular}
\caption{Top 5-gram LSTMs.}
\label{t:ngram_lstm}
\end{table}

\section*{Results by pattern}
See Table \ref{t:hebrew} and Table \ref{t:russian} for Hebrew and
Russian, respectively (Italian and English results are available in
the main paper).
\begin{table}[p]
\centering
\begin{tabular}{lrrrrr}
\toprule
construction & acc original & \textit{std} & acc  nonce & \textit{std} & size\\
\midrule
NOUN ADJ PROPN VERB & 100.0 & 0.0 & 61.8 & 9.5 & 50\tabularnewline
NOUN NOUN ADJ VERB & 100.0 & 0.0 & 90.7 & 3.5 & 60\tabularnewline
NOUN NOUN PROPN VERB & 100.0 & 0.0 & 80.0 & 4.2 & 30\tabularnewline
NOUN PROPN VERB & 100.0 & 0.0 & 100.0 & 0.0 & 10\tabularnewline
NOUN VERB PUNCT VERB & 95.0 & 11.2 & 81.7 & 4.2 & 40\tabularnewline
NOUN VERB VERB & 86.7 & 9.3 & 65.7 & 4.1 & 90\tabularnewline
NOUN ADJ DET ADJ & 99.4 & 0.8 & 96.4 & 1.0 & 710\tabularnewline
NOUN ADJ PUNCT SCONJ VERB & 98.8 & 2.8 & 89.6 & 1.8 & 160\tabularnewline
NOUN DET ADV PUNCT ADJ & 93.3 & 5.3 & 81.2 & 4.6 & 300\tabularnewline
NOUN NOUN DET ADJ & 94.2 & 1.0 & 77.6 & 0.9 & 1520\tabularnewline
NOUN NOUN PUNCT SCONJ VERB & 90.5 & 6.7 & 73.9 & 2.7 & 210\tabularnewline
NOUN PUNCT NOUN CCONJ NOUN & 96.7 & 4.6 & 50.7 & 3.7 & 120\tabularnewline
NOUN PUNCT NOUN PUNCT NOUN & 98.3 & 3.7 & 68.3 & 4.1 & 120\tabularnewline
VERB NOUN CCONJ VERB & 83.3 & 5.9 & 83.1 & 2.8 & 240\tabularnewline
VERB NOUN NOUN CCONJ VERB & 100.0 & 0.0 & 80.0 & 9.3 & 10\tabularnewline
VERB NOUN PUNCT CCONJ VERB & 100.0 & 0.0 & 73.9 & 5.4 & 40\tabularnewline
VERB VERB CCONJ VERB & 90.0 & 22.4 & 73.3 & 9.1 & 20\tabularnewline
\bottomrule
\end{tabular}
\caption{Average top 5 LSTM accuracies for Hebrew.}
\label{t:hebrew}

\end{table}

\begin{table}[p]
\centering
\begin{tabular}{lrrrrr}
\toprule
construction & acc original & \textit{std} & acc  nonce & \textit{std} & size\\
\midrule
ADJ ADJ NOUN & 99.1 & 0.9 & 94.1 & 0.9 & 910\tabularnewline
ADJ DET NOUN & 100.0 & 0.0 & 96.9 & 0.5 & 370\tabularnewline
ADJ NOUN ADJ NOUN & 97.3 & 3.7 & 96.6 & 1.5 & 150\tabularnewline
ADJ NOUN NOUN & 88.6 & 2.6 & 69.6 & 2.2 & 420\tabularnewline
ADJ PUNCT ADJ NOUN & 99.4 & 0.9 & 94.5 & 0.8 & 620\tabularnewline
ADJ PUNCT NOUN NOUN & 72.0 & 11.0 & 71.1 & 4.4 & 50\tabularnewline
ADJ PUNCT VERB NOUN & 80.0 & 18.3 & 84.4 & 3.1 & 30\tabularnewline
ADJ VERB NOUN & 80.0 & 11.2 & 83.3 & 5.9 & 40\tabularnewline
DET ADJ ADJ NOUN & 100.0 & 0.0 & 99.2 & 1.1 & 190\tabularnewline
DET ADJ NOUN & 97.1 & 2.6 & 94.6 & 1.8 & 210\tabularnewline
DET VERB NOUN & 97.5 & 5.6 & 75.8 & 5.8 & 80\tabularnewline
NOUN PUNCT VERB ADV VERB & 100.0 & 0.0 & 85.6 & 3.0 & 40\tabularnewline
VERB ADJ NOUN & 94.0 & 5.5 & 86.7 & 2.2 & 100\tabularnewline
VERB NOUN ADJ NOUN & 100.0 & 0.0 & 98.5 & 1.6 & 90\tabularnewline
VERB NOUN NOUN & 87.0 & 5.3 & 66.7 & 3.4 & 230\tabularnewline
NOUN PUNCT CCONJ PUNCT NOUN & 93.3 & 14.9 & 83.7 & 7.2 & 30\tabularnewline
VERB NOUN CCONJ PART VERB & 98.2 & 4.1 & 91.7 & 2.3 & 110\tabularnewline
VERB NOUN CCONJ VERB & 95.2 & 1.9 & 86.7 & 1.6 & 670\tabularnewline
VERB NOUN NOUN CCONJ VERB & 100.0 & 0.0 & 96.1 & 1.5 & 40\tabularnewline
VERB PROPN CCONJ VERB & 100.0 & 0.0 & 93.9 & 4.6 & 40\tabularnewline
\bottomrule
\end{tabular}
\caption{Average top 5 LSTM accuracies for Russian.}
\label{t:russian}
\end{table}

\section*{Human data}

Subjects were recruited through the standard Amazon Mechanical Turk
interface.\footnote{\url{https://www.mturk.com/}} Instructions were
presented in Italian, and subjects were paid 2 dollar cents per
assignment. We did not record personal data about subjects, and our
data set only contains aggregated data.

%\subsection{Availability of source code and data}

% We made the scripts allowing to reproduce our results available on a
% github repository that is currently private to protect anonymity.
% This includes scripts to generate our language modeling training data
% and train our language models, as well as scripts to generate our
% long-distance agreement test sets and carry out our evaluation. Upon
% publication, code and data will be available under the free license that
% will be recommended by our institution based on internal review. We uploaded the long-distance
% agreement test sets together with our submission for reviewers perusal
% only.